\documentclass[sn-standardnature,iicol]{sn-jnl}% Standard Nature Portfolio Reference Style
\usepackage{graphicx}
\usepackage{subcaption}
\usepackage{float}
\usepackage{amsmath}
\jyear{2021}%

\theoremstyle{thmstyleone}%
\theoremstyle{thmstyletwo}%

\theoremstyle{thmstylethree}%

\begin{document}

\title[Exploring the Trie of Rules]{Exploring the Trie of Rules: a fast data structure for the representation of association rules}


\author*[1,2]{\fnm{Mikhail} \sur{Kudriavtsev}}\email{mikhail.kudriavtsev2@mail.dcu.ie}

\author[2,3]{\pfx{Dr} \fnm{Marija} \sur{Bezbradica}}\email{marija.bezbradica@dcu.ie}

\author[1,2]{\pfx{Dr} \fnm{Andrew} \sur{McCarren}}\email{andrew.mccarren@dcu.ie}

\affil*[1]{\orgdiv{Centre for research training in artificial intelligence
(CRT-AI)}, \orgname{Dublin City University}, \orgaddress{ \city{Dublin}, \country{Ireland}}}

\affil[2]{\orgdiv{Adapt Research Centre}, \orgname{Dublin City University}, \orgaddress{ \city{Dublin}, \country{Ireland}}}

\affil[3]{\orgdiv{Insight Centre for Data Analytics}, \orgname{Dublin City University}, \orgaddress{ \city{Dublin}, \country{Ireland}}}

\abstract{
    %intro
Association rule mining techniques can generate a large volume of sequential data when implemented on transactional databases. Extracting insights from a large set of association rules has been found to be a challenging process. When examining a ruleset, the fundamental question is how to summarise and represent meaningful mined knowledge efficiently. Many algorithms and strategies have been developed to address issue of knowledge extraction; however, the  effectiveness of this process can be limited  by the data structures.  A better data structure can sufficiently affect the speed of the knowledge extraction process.\\
 %specifics
 This paper proposes a novel data structure, called the Trie of rules, for storing a ruleset that is generated by association rule mining. The resulting data structure is \textit{a prefix-tree graph structure made of pre-mined rules}. This graph stores the rules as paths within the prefix-tree in a way that similar rules overlay each other. Each node in the tree represents a rule where a consequent is this node, and an antecedent is a path from this node to the root of the tree.\\
%findings
 The evaluation showed that the proposed representation technique is promising. It compresses a ruleset with almost no data loss and benefits in terms of time for basic operations such as searching for a specific rule and sorting, which is the base for many knowledge discovery methods. Moreover, our method demonstrated a significant improvement in traversing time, achieving an 8-fold increase compared to traditional data structures.

}




\maketitle

\section{Introduction}\label{sec1}
Modern organizations produce, gather, and store substantial volumes of data with the intention of using it to make strategic decisions. The level of complexity in making these decisions can often be exasperated by the depth of understanding required to interpret relationships within the data. Retailers are a typical example, as they utilize regular sequences or patterns in their customer transactions in order to make marketing or supply chain management strategic decisions.

Association Rule Mining (ARM) is a data mining technique that extracts frequent patterns and relationships among items in a dataset. The process of ARM involves obtaining the dataset, applying an appropriate frequent item mining algorithm to it, and extracting association rules from the frequent itemsets generated by the algorithm. These association rules are stored in a data structure called a \textit{ruleset}, which can be very large and computationally expensive to process. The data structure used for a ruleset is critical, as it determines the efficiency and effectiveness of knowledge extraction methods that are based on traversing the ruleset~\cite{Alasow2020, Bui-Thi2020, Li2014}.

Formally, a rule can be described as an implication $A \rightarrow C$ where $A$ is the antecedent, and $C$ is the consequent, $A$ and $C$ are sets of items from $I={i_1,i_2,...,i_m}$ where $A \cap C = \emptyset $. Each rule is derived from a database $D = {t_1,t_2,...,t_n}$, where each transaction $t_i$ is a set of items from $I$.  

Association rule mining algorithms were introduced by Agrawal~\cite{Agrawal} to identify the buying patterns of retail customers.  These techniques are based on the exploration of frequently occurring sequences in large-scale databases and have since been widely used in many application areas such as customer behavior analysis, software engineering, medical diagnostics, visual analytics, etc.~\cite{Yazgana2016, Shaukat2017, Ghafari2019,  7192697}. Generally, the output of an ARM algorithm is a list of rules or \textit{association ruleset} portrayed in a text-based manner, requiring further analysis. Rules have typically been assessed using evaluation metrics such as Support, Confidence, and Lift~\cite{Bayardo1999}.
 
%\Difference between a frequent sequence and an association rule
The major difference between an \textit{association rule} and a \textit{frequent sequence} should be emphasized because it is vital for the future explanation of the proposed methodology. A frequent sequence is a list of items that appears as an output of an ARM algorithm. An association rule is a structure made from the frequent sequence by splitting this sequence into an antecedent and consequent. For example, a frequent sequence $(a, b, c, d)$ is mined with an ARM algorithm; a rule $(a, b) \rightarrow (c, d)$ can be constructed using this sequence, where $(a, b)$ is the antecedent and $(c, d)$ is the consequent. As a result of that difference, evaluation metrics applied to association rules can not be used for frequent sequences~\cite{Agrawal, Brin1997}. The only exception is Support~\cite{Agrawal}; it reflects the frequency of all items together in a rule or a sequence.

While a considerable amount of attention has been directed toward data structures for \textit{frequent sequence sets} in the context of ARM~\cite{Bodon2003, Grahne2003a, Coenen2004},  there has been limited focus on identifying a suitable data structure that allows efficient knowledge extraction and  storing \textit{sets of association rules}. Relatively less attention has been given to exploring how to store such data structures and utilize them for further data exploration and use with other ML methods~\cite{Li2014, Bui-Thi2021}. As ARM continues to remain a popular field of study, identifying a data structure that can effectively store and represent association rules would  improve current knowledge extraction in state-of-the-art approaches.

The FP-tree is a widely used data structure for storing frequent sequences because it is an efficient data structure for storing sequence-based data. However, it has not been properly explored as an application for storing a ruleset.

In this paper, we present a novel use of the FP-tree for storing a ruleset, rather than a frequent sequence set, which is its typical use case. By doing so, we improve the efficiency and effectiveness of  knowledge extraction methods that are based on traversing the ruleset.

\section{Background}\label{sec2}  

\begin{figure*}
  \centering
  \includegraphics[width=\textwidth]{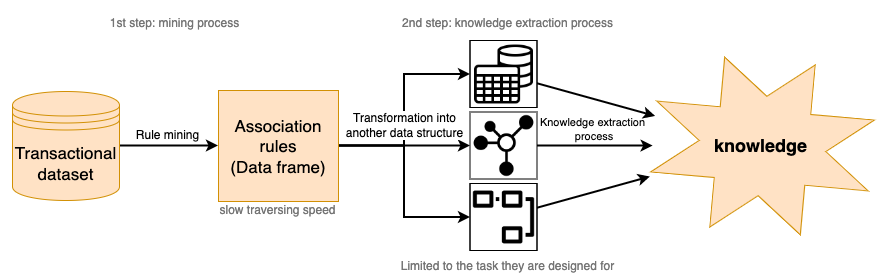}
  \caption{The process of ARM knowledge discovery}
  \label{fig:ARM_pipeline}
\end{figure*}

 Due to the complex nature of association rules, the use of data structures plays a critical role in Association Rule Mining techniques. Association rules consist of two parts, and the order of these parts is crucial, with each part potentially containing multiple components. Therefore, proper representation of rules can significantly impact the efficiency of ARM processes. The overall process of ARM can be broadly divided into two phases~(Fig.~\ref{fig:ARM_pipeline}): the mining process, where all rules are discovered, and the knowledge extraction process which involves the evaluation of rules, sorting, filtering, visualization, grouping, etc. In both phases, the choice of data structure directly impacts the efficiency and effectiveness of the ARM algorithms. Selecting appropriate data structures for ARM is essential for ensuring the accuracy and efficiency of the mining and knowledge extraction processes~\cite{Alasow2020, Bui-Thi2020, Li2014}.

% insert pipline justification of what is where

 One of the key aspects of mining association rules is the efficient storage and retrieval of frequent itemsets, which are the building blocks for generating rules.  ARM algorithms utilize various data structures for the mining process, such as the Apriori algorithm that uses a hash tree or hash table to store frequent itemsets in a compact manner to reduce the memory overhead~\cite{Coenen2004, Agrawal_apriori}. Another popular approach is the FP-growth algorithm that employs a data structure called FP-tree or trie, which allows for efficient mining of frequent itemsets by compressing the transactions into a compact tree-like structure~\cite{Vu2011,Grahne2003a}. ECLAT (Equivalence Class Transformation) algorithm uses a vertical data structure, where transactions are sorted by items, and each item is associated with a list of transactions it appears in, facilitating the generation of association rules through traversal of this data structure~\cite{Zaki1997}.  These data structures enable the mining algorithms to quickly identify frequent itemsets and generate association rules from large transactional datasets.

\subsection{Data structures in knowledge extraction process}
While the topic of data structures for ARM \textit{mining process} is well discussed in the literature, there is a notable lack of discussion on the data structures for \textit{knowledge extraction process}. The main focus has been on how to obtain knowledge, with data structures being discussed only as a tool, rather than as an essential aspect of the knowledge extraction process itself. Meanwhile, a ruleset can be extensive, making it challenging to efficiently explore and assess rules to derive valuable knowledge from it. Furthermore, there is no universal agreement on the right data structure for representing a ruleset,  and many widely used libraries for ARM, such as mlxtend~\cite{Stancin2019} and libraries by M. Hahsler~\cite{Hahsler2023}, output the ruleset as a data frame, which is essentially a plain table. However, this simplistic data structure is inefficient for performing knowledge extraction tasks on large rulesets. As a result, various knowledge extraction methods have to transform a ruleset into a manageable data structure.

% other methods are limited
In the field of ARM, many techniques are employed to extract knowledge by transforming the ruleset into a graph-like data structure. This approach has proven to be efficient for representing and manipulating rules, as cited by various studies~\cite{Yen2019, Koh2010, Hahsler2016, Jentner2019a, DePadua2018, Berrado2007}. However, these methods are tailored to specific applications and may not be suitable for other tasks. For example, Bui-Thi et al.~\cite{Bui-Thi2020} convert a ruleset into a feature vector, which is advantageous for clustering but inconvenient for other tasks like traversing the data structure or obtaining rule metrics due to data loss. This trend has been observed multiple times in the literature.  

% dataframe is slow
Many knowledge extraction methods still use traditional data frames to represent the ruleset~\cite{Hahsler2023, Buchta2022}, which may not be as efficient for large-scale rule sets. For example, a common approach is to represent the ruleset as a plain table or a matrix, where each row corresponds to a rule, and each column represents different attributes of the rule, such as the antecedent, consequent, Support, Confidence, and Lift. However, this approach can become computationally expensive and memory-consuming when dealing with a large number of rules, which may limit the scalability and performance of the knowledge extraction process~\cite{Moahmmed2021}.
\begin{figure*} 
    \centering
     \includegraphics[width=\textwidth]{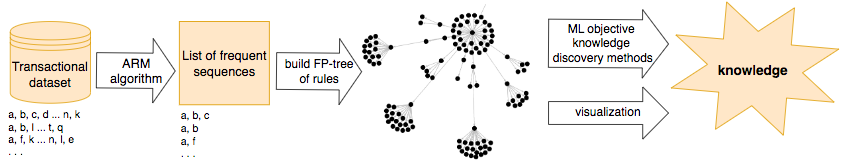}
      \caption{Proposed data representation process.}
      \label{fig:visualisation_pipeline_tor}
\end{figure*}

Another widely used data structure in association rule mining is the FP-tree (Frequent Pattern tree), which was originally proposed for efficient mining of \textit{frequent itemsets} in transactional databases~\cite{Bodon2003}. The FP-tree represents the frequent itemsets as a compact tree structure, allowing for efficient storage and retrieval of frequent patterns. However, the use of FP-tree as a data structure for association rules is relatively limited and mainly focused on classification rules, rather than association rules~\cite{Li2001, AzevedoAClassifiers}. For instance, FP-viz, a visualization tool that utilizes the FP-tree structure, has been proposed for visualizing frequent sequences but not association rules~\cite{Keim2005}. However, further research and exploration are needed to investigate the potential of FP-tree as a data structure for storing and manipulating association rules.

\subsection{Evaluation metrics}
Evaluation of association rules is a crucial step in ARM knowledge extraction process. More than 40 metrics can be utilized for assessing an association rule~\cite{arules:Geng:2006, Wu2010, Luna2018}, with Support, Confidence, and Lift being among the most widely used.\\

$\text{Support}(X \Rightarrow Y) = \frac{\text{Transactions X and Y}}{\text{Total number of transactions}}$\\

$\text{Confidence}(X \Rightarrow Y) = \frac{\text{Support}(X \cup Y)}{\text{Support}(X)}$\\

$\text{Lift}(X \Rightarrow Y) = \frac{\text{Confidence}(X \Rightarrow Y)}{\text{Support}(Y)}$\\

In the formulas above, $X$ and $Y$ represent the antecedent and consequent of the association rule, respectively. Support measures the frequency of occurrence of the rule in the dataset, Confidence measures the conditional probability of $Y$ given $X$, and Lift measures the degree of association between $X$ and $Y$, taking into account the expected Support of $Y$.

As seen from the formulas, the Support metric does not differentiate between the antecedent and consequent, while  Confidence and Lift take into account the relationship between them. This implies that the evaluation metrics used in ARM impose certain constraints on the choice of data structures that can be employed for storing, representing, and processing association rules. For example, FP-viz, a visualization tool, can only display Support and is therefore suitable for assessing only frequent sequences~\cite{Keim2005}. Similarly, Azevedo et al. attempted to use an FP-tree-like structure for association rules, but faced limitations in handling rules with consequents consisting of more than one item, making it applicable only for classification model~\cite{AzevedoAClassifiers}.

 In summary, this section highlights the importance of choosing appropriate data structures for representing association rules in the ARM knowledge extraction process. While various data structures have been proposed and used, the dataframe remains the primary data structure due to its versatility and transformability. Nevertheless, it is not without limitations, including slow performance, large size, and occasional redundancy. Therefore, there is a need for a robust data structure that can perform simple tasks such as traversing, searching, filtering, accessing metrics, and be used for sophisticated knowledge extraction methods. In addition, ideally, it should be quick and easy to interpret, visualize, explore, and understand.

\section{Methodology}\label{sec3}

 The popular FP-tree data structure, commonly used for storing frequent sequences~\cite{Bodon2003, Han2004, Grahne2003a}, has not been extensively explored as a data structure for association rules. Therefore, in this study, we propose a novel use of this data structure, called the "Trie of rules", for representing association rules that can serve as a viable alternative to a regular dataframe. Our proposed data structure takes the form of a graph that contains all the rules, associated metric values, and avoids redundancy, while also increasing the traverse speed. This data structure is suitable for various graph-based knowledge extraction methods and vizualization, as well as for efficient storage and retrieval of rules. Fig.~\ref{fig:visualisation_pipeline_tor} shows the ARM pipeline incorporating the Trie of rules. The best way to explain the concept is through a three-step process:

\begin{enumerate}
    \item \textbf{Step 1}: apply an ARM algorithm on a transactional dataset to acquire a list of frequent sequences.
    
    \item \textbf{Step 2}: build an FP-tree using the list of frequent sequences from Step~1. Each node in the FP-tree represents a rule. The rule is a path from the root to a desired node, where a consequent is the last node in the path, and an antecedent is all the previous nodes in the same path before the desired node. (Fig.~\ref{fig:rule_structure}). 
   
    \item \textbf{Step 3}: label each node  with the Support, Confidence, Lift, and other values relating to the corresponding rule~(Fig.~\ref{fig:rule_structure}). 

\begin{figure}[H]
     \centering
     \includegraphics[width=2.9in]{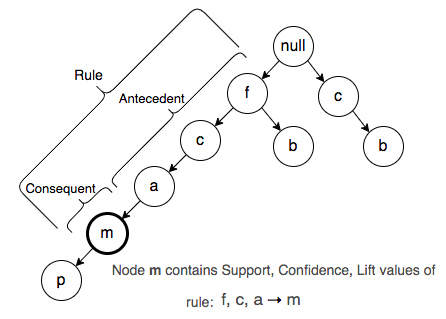}
     \caption{The structure of a rule in a Trie of rules.}
      \label{fig:rule_structure}
\end{figure}

\end{enumerate}

\subsection{An illustrative example} \label{pilot_example} 
Let's take an example from a simple dataset to demonstrate the advantages of using the Trie of rules approach. The analysis will be done in detail.

\begin{enumerate}
    \item \textbf{Step 1}: In the first step an FP-max algorithm is applied to the dataset in~Fig.~\ref{fig:dataset_pilot_example} to acquire a list of frequent sequences. The chosen minimum Support threshold value for this dataset is $0.3$. It should be noted here that the choice of ARM algorithm is user specific. In this example the FP-max algorithm is used because it usually produces a smaller output volume. The FP-max algorithm resulted in the creation of  a table with frequency of each individual items (Fig.~\ref{fig:table_items_frequency}) and three frequent sequences: ${(f, c, a, m, p)}$, ${(f, b)}$, ${(c, b)}$ (Fig.~\ref{fig:table_frequent_sequences}).
    \begin{figure}[H]
    \centering
    \subfloat[]{\includegraphics[width=1.2in]{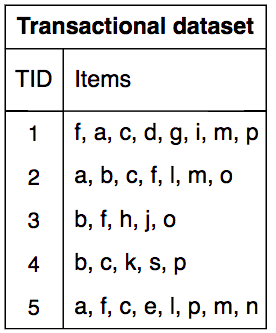}
    \label{fig:dataset_pilot_example}}
    \hfil
    \subfloat[]{\includegraphics[width=0.9in]{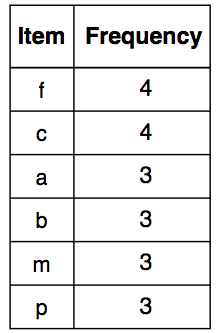}
    \label{fig:table_items_frequency}}
    \hfil
    \subfloat[]{\includegraphics[width=1.2in]{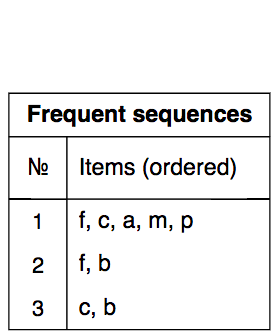}
    \label{fig:table_frequent_sequences}}
    \caption{Dataset~(a) and Step~1 outcomes: (b) frequent items and (c) frequent sequences. }
    \end{figure}
    
    \item \textbf{Step 2}:
    The list of frequent sequences, generated in the previous step, serves as a source dataset for the creation of an FP-tree.
    The FP-tree is initialized with a root node (Null), followed by inserting the \textit{frequent sequences} one by one into the FP-tree. Before the insertion, items in each frequent sequence are sorted \textit{according to their frequency in the original dataset}.  In this example, the first sequence inserted is ${(f, c, a, m, p)}$;  The items of the sequence are inserted into the FP-tree as nodes. Fig.~\ref{fig:trie_1} shows the FP-Tree when the first sequence has been fully traversed. Following this, the second sequence  ${(f, b)}$ is inserted into the FP-tree. Note that the $f$ element  has occurred beforehand, which can also be seen in the FP-Tree created so far. Therefore, instead of creating a new branch, the second sequence overlays the existing FP-tree and creates an additional branch only when $b$ item occurs. After traversing the second sequence, the FP-Tree looks as shown in~Fig.~\ref{fig:trie_2}. The last sequence to be inserted is $(c, b)$. Since this sequence differs from others in terms of its first item, a new branch is created from the root. Finally, after traversing all the sequences, the FP-Tree appears as shown in~Fig.~\ref{fig:trie_3}. 
    \begin{figure}[H]
    \centering
    \subfloat[]{\includegraphics[width=1.5in]{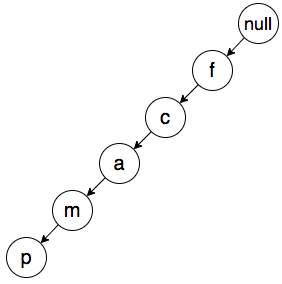}
    \label{fig:trie_1}}
    \hfil
    \subfloat[]{\includegraphics[width=1.5in]{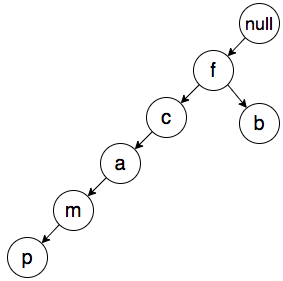}
    \label{fig:trie_2}}
    \\
    \subfloat[]{\includegraphics[width=2in]{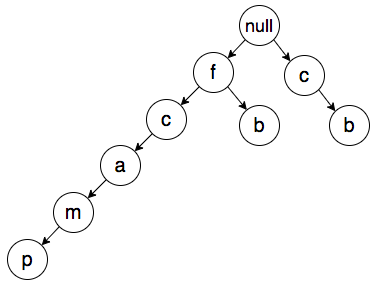}
    \label{fig:trie_3}}
    \caption{Step 2 process. The Trie of rules after: the first~(a), second~(b), and third~(c) frequent sequences are inserted.}
    \end{figure}

 \item \textbf{Step 3}:   Each node in the FP-tree is extended with metrics such as  Support, Confidence, Lift, and others, corresponding to the rule that this node represents, Fig.~\ref{fig:metrics_insertion}. 
 \begin{figure}[H]
     \centering
     \includegraphics[width=2.5in]{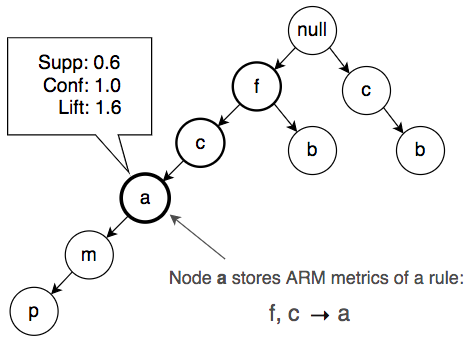}
     \caption{Step 3. ARM metrics of node \textit{a}.}
      \label{fig:metrics_insertion}
\end{figure}
    
 \end{enumerate}

\subsection{Confidence calculation for a compound consequent} \label{confidence}

\begin{figure}[!h]
 \centering
\centerline{
\includegraphics[width=2.8in]{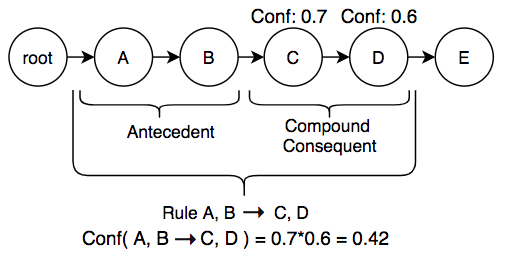}}
\caption{A rule with a compound consequent.}
\label{fig:compound_consequent}
\end{figure}

In a Trie of rules each node shows Confidence only for a rule with a single-item consequent; however, the proposed representation model can be used to derive the value of Confidence for more complex rules directly from the graph. The Confidence of a compound-consequent rule can be calculated as multiplication of Confidence values of the nodes in the consequent (Fig.~\ref{fig:compound_consequent}). % the nodes in the compound consequent should be in a sequence without skips and interruptions. 
 
This feature is possible because of the specifics of a Trie of rules. In order to calculate Confidence of a rule two values are used: Support of the antecedent and Support of the whole rule.
 As mentioned, the Trie of rules is based on the FP-tree structure. Hence, every path starting from the root is unique because identical sequences will overlay each other in one path. Consequently, when picking a node and observing a Support value in it, one can be sure that this value represents true Support for the sequence equal to the path to this node. Therefore, the calculation of Confidence does not require any information from other branches. All this allows to multiply Confidence values for a sequence of nodes which further allows to evaluate rules with compound consequent in a Trie of rules.  A multiplication formula of Confidence can be derived as follows: 

\begin{equation}
Conf(A,B \rightarrow C,D)=\frac{Sup(A,B,C,D)}{Sup(A,B)}
\label{eq:equation1}
\end{equation}

\begin{equation}
Conf(A,B \rightarrow C)=\frac{Sup(A,B,C)}{Sup(A,B)}
\label{eq:equation2}
\end{equation}

\begin{equation}
Conf(A,B,C \rightarrow D)=\frac{Sup(A,B,C,D)}{Sup(A,B,C)}
\label{eq:equation3}
\end{equation}
To prove the point, Equation \ref{eq:equation1} can be derived by combining Equation \ref{eq:equation2} and Equation \ref{eq:equation3}.

 \begin{multline}
Conf(A,B \rightarrow C ) * Conf(A,B,C \rightarrow D ) = \cr  \frac{Sup(A,B,C)}{Sup(A,B)} * \frac{Sup(A,B,C,D)}{Sup(A,B,C)} = \cr\frac{Sup(A,B,C,D)}{Sup(A,B)} = Conf(A,B \rightarrow C,D )
 \end{multline}

\subsection{Discussion on methodology}

Initially, it may appear that not all rules are included in the presented graph due to the usage of frequent sequences for insertion. However, it is important to note that the data structure employed in our methodology ensures that all items within each inserted rule are sorted based on their frequency in descending order. This sorting approach allows us to avoid false Confidence situations~\cite{Brin1997} and focus solely on the most valuable rules. Furthermore, the information contained within this data structure enables the derivation of additional rules, as it encompasses all items along with their corresponding Support values. Thus, our approach not only captures the most significant rules but also provides a foundation for extracting further insights from the data.

The next section presents an evaluation of the proposed data structure.

\section{Evaluation}\label{sec4}
The evaluation of the proposed data structure is crucial in determining its usefulness in further use with knowledge discovery methods. In this section, we present a comparison of the proposed method with the popular in the field data structure for a ruleset used in various Python libraries, which is the \textit{Pandas data frame}~\cite{Hahsler2023, Buchta2022}, and describe how our method affects basic operations in terms of time efficiency. Specifically, we analyze the time taken to perform the following operations:
\begin{itemize}
    \item finding a rule and its metrics in a ruleset,
    \item the average time to find a rule considering the size of a ruleset,
    \item time taken to create a set of rules from a set of transactions and its dependence on the minimum threshold value,
    \item time taken to get the top N rules by Support from a ruleset,
    \item time taken to get the top N rules by Confidence from a ruleset.
\end{itemize}
By conducting these evaluations, we aim to demonstrate the effectiveness and efficiency of our proposed data structure for association rule mining.

To assess the proposed data structure, we use a grocery dataset found in the R Project for Statistical Computing package "arules"~\cite{Buchta2022}. The dataset includes data on $9834$ transactions and $169$ unique items. A minimum Support value of $0.005$ is empirically chosen, which results in approximately $1000$ frequent sequences and $3000$ association rules using the Apriori algorithm.
 
\begin{figure}[H]
     \centering
     \includegraphics[width=2.9in]{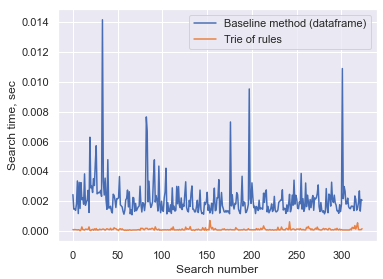}
     \caption{Comparison of search time of rules in a ruleset between Trie of Rules and Pandas Data Frame}
      \label{fig:search_time}
\end{figure}

\begin{figure}[H]
     \centering
     \includegraphics[width=2.9in]{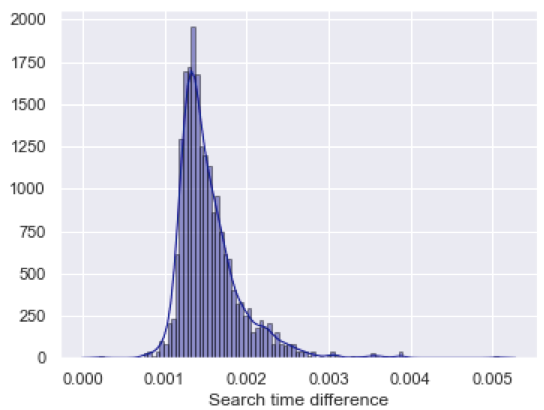}
     \caption{Distribution of differences between search time in Trie of rules and Pandas data frame}
      \label{fig:ttest}
\end{figure}

 The search operation for finding a rule and its metrics in a ruleset is fundamental and can be considered vital in exploring a dataset, as it is a task of random access to data. Many operations, from searching to clustering, are based on this operation. To evaluate the performance of the Trie of rules in comparison with the Pandas data frame, we conducted an experiment where every rule was searched in both data structures. The average time taken for the Trie of rules was found to be $0.000146$ seconds, while for the Pandas data frame it was $0.00123$, demonstrating that our method outperforms the baseline methodology on average by eight times~(Fig.~\ref{fig:search_time}). To establish the significance of this difference, a t-test was performed with a null hypothesis that the difference in times between these methods is zero, with an alternative hypothesis that the difference is not zero~(Fig.~\ref{fig:ttest}). The null hypothesis was rejected with a p-value of $1e-245$ confirming that the difference is statistically significant.

\begin{figure}[H]
     \centering
     \includegraphics[width=2.9in]{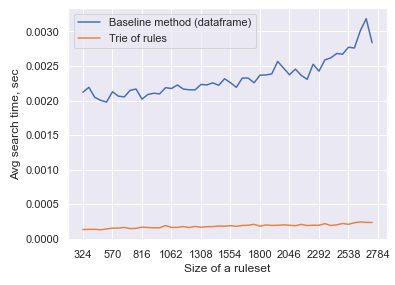}
     \caption{Comparison of average time to search a rule in a set of rules with various minimum Support values in the range between 0.005 to 0.0135 (a lower minimum Support value corresponds to a larger number of rules in the set)}
      \label{fig:search_size}
\end{figure}

Furthermore, a series of tests were conducted with different minimum-Support thresholds to examine how the size of a ruleset affects the search time. As depicted in Fig.~\ref{fig:search_size}, the search time for Trie of rules continues to outperform Pandas Data Frame as the number of rules increases.

\begin{figure}[H]
     \centering
     \includegraphics[width=2.9in]{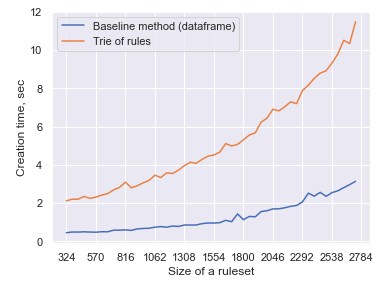}
     \caption{Comparing ruleset creation time with different Support values.
     (a lower minimum Support value corresponds to a larger number of rules in the set)
     }
      \label{fig:create_time}
\end{figure}

 However, one of Trie of rules limitations is its construction time. Fig.~\ref{fig:create_time} demonstrates the experimental results and indicates that as the minimum Support threshold decreases (and the size of the ruleset consequently increases), the construction time for the data structure also increases.
 
\begin{figure}[!th]
\centering
\subfloat[]{\includegraphics[width=2.9in]{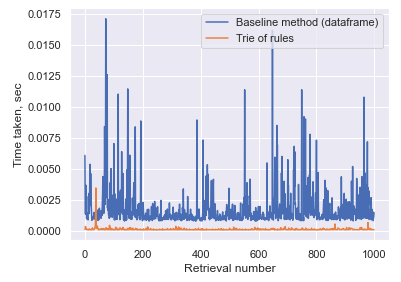} }
\hfil
\subfloat[]{\includegraphics[width=2.9in]{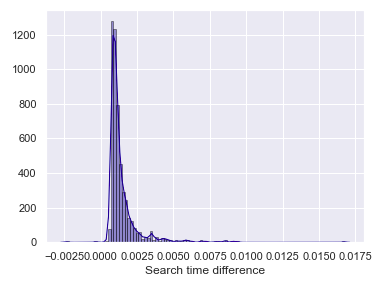} }
\caption{Experimental results of measuring the time taken to retrieve top 10\% rules by Support in Trie of rules and Pandas dataFrame. (a) Comparison of retrieval time. (b) Distribution of differences between retrieval times.}
\label{fig:support_sort}
\end{figure}

\begin{figure}[!th]
\centering
\subfloat[]{\includegraphics[width=2.9in]{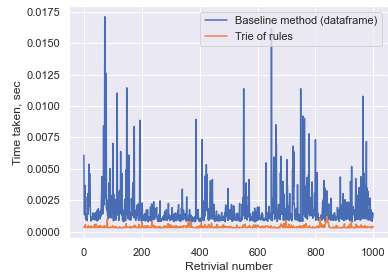} }
\hfil
\subfloat[]{\includegraphics[width=2.9in]{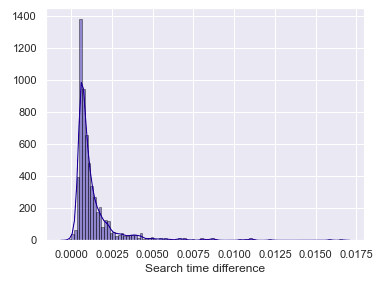} }
\caption{Experimental results of measuring the time taken to retrieve top 10\% rules by Confidence in Trie of rules and Pandas dataFrame. (a) Comparison of retrieval time. (b) Distribution of differences between retrieval times.}
\label{fig:conf_sort}
\end{figure}

In order to investigate the impact of search time on other operations, two experiments were conducted to assess the efficiency of the Trie of rules data structure in retrieving top N rules from a ruleset. Specifically,  we measured the time required to retrieve the top $10\%$ rules based on Support and the top $10\%$ rules based on Confidence. The results, as shown in Fig.~\ref{fig:support_sort} and Fig.~\ref{fig:conf_sort}, indicate that the Trie of rules outperforms the typical data frame, as confirmed by a t-test with rejected null hypotheses that  the mean of differences is equal to $0$ and p-value $<0.05$ for both experiments.  

% \begin{figure}[H]
%      \centering
%      \includegraphics[width=2.9in]{5img_evaluation_supp_sort.png}
%      \caption{!!!Comparison of time to create a set of rules with various minimum support values in the range between 0.005 to 0.0135 ( a lower minimum support value corresponds to a larger number of rules in the set)}
%       \label{fig:support_sort}
% \end{figure}

Similar experimental results were obtained using a larger dataset of online retail logs available at~\cite{retail}. This dataset is more sparse, consisting of approximately $18,000$ transactions with $3,600$ different items. A minimum Support threshold of $0.002$ was chosen, resulting in approximately $45,000$ frequent sequences and $300,000$ association rules, making the ruleset large enough for evaluation. The mining time for the Trie of rules was approximately 25 minutes, whereas creating a DataFrame with rules took only 2 minutes. However, traversing through all rules in the Trie of rules took only 25 minutes, while for Pandas DataFrame it took more than 2 hours. It's important to note that  all the experiments were conducted on the same machine within the same environment. The main conclusion is that, despite taking more time to construct, the Trie of rules is a fast data structure. Given that creating a ruleset is typically a one-time task, this data structure might increase productivity in knowledge extraction by allowing for faster traversal and exploration of the ruleset.

\section{Conclusion}\label{sec5}
Association rule mining is a popular method used for knowledge discovery in various fields. However, little attention has been given to the data structure used to store the resulting ruleset. In this paper, we proposed the Trie of rules\footnote{https://github.com/ARM-interpretation/Trie-of-rules} data structure for storing a ruleset, which can be utilized for further investigation using machine learning methods.

The proposed Trie of rules data structure has been shown to significantly improve the time efficiency of knowledge discovery methods in association rule mining. By organizing rules into a prefix-tree graph structure, the proposed data structure enables faster traversal through the ruleset, reducing the time complexity of knowledge discovery methods. Furthermore, this data structure has the potential to enhance the accuracy and comprehensibility of rule-based systems, as it potentially provides a comprehensive visualization structure that eases the subjective exploration process.

While the Trie of rules shows promising results, further investigation is needed to research the space efficiency and increase mining efficiency of this method.  Overall, the Trie of rules data structure provides a useful tool for knowledge extraction, and it can contribute to the development of more efficient and effective approaches for ARM.

\bmhead{Acknowledgments}
This publication has emanated from research conducted with the financial Support of Science Foundation Ireland under Grant number 18/CRT/6223 For the purpose of Open Access, the author has applied a CC BY public copyright license to any Author Accepted Manuscript version arising from this submission.

\bmhead{Authors contribution statement}
Mikhail Kudriavtsev led and executed the main work, including idea development, writing, and conducting experiments. Dr Andrew McCarren served as the principal supervisor. Both Dr Andrew McCarren  and Dr Marija Bezbradica contributed to the writing process and the development of the approach.

\bmhead{Competing Interests}
The authors declare that they have no known competing financial interests or personal relationships
that could have appeared to influence the work reported in this paper.

\bmhead{Data availability and ethical and informed consent for data used}
The data used in this paper is freely available in publicly accessible repositories~\cite{retail,Buchta2022}, and its usage adheres to all ethical guidelines and regulations. No ethical issues are associated with the use of this data in our study.

\bibliography{references,non_mendeley_refences}

\begin{thebibliography}{10}
\expandafter\ifx\csname url\endcsname\relax
  \def\url#1{\burl{#1}}\fi
\expandafter\ifx\csname urlprefix\endcsname\relax\def\urlprefix{URL }\fi
\providecommand{\bibinfo}[2]{#2}
\providecommand{\eprint}[2][]{\url{#2}}
\providecommand{\doi}[1]{\url{https://doi.org/#1}}
\bibcommenthead

\bibitem{Alasow2020}
\bibinfo{author}{Alasow, M.~A.}, \bibinfo{author}{Mohammed, S.~A.} \&
  \bibinfo{author}{El-Alfy, E. S.~M.}
\newblock \bibinfo{title}{{Parallel Association Rules Pruning Algorithm on
  Hadoop MapReduce}} (\bibinfo{year}{2020}).
\newblock
  \urlprefix\url{https://link.springer.com/chapter/10.1007/978-981-15-3852-0_8}.

\bibitem{Bui-Thi2020}
\bibinfo{author}{Bui-Thi, D.}, \bibinfo{author}{Meysman, P.} \&
  \bibinfo{author}{Laukens, K.}
\newblock \bibinfo{title}{{Clustering association rules to build beliefs and
  discover unexpected patterns}}.
\newblock \emph{\bibinfo{journal}{Applied Intelligence}}
  \textbf{\bibinfo{volume}{50}}~(6), \bibinfo{pages}{1943--1954}
  (\bibinfo{year}{2020}).
\newblock \doi{10.1007/s10489-020-01651-1} .

\bibitem{Li2014}
\bibinfo{author}{Li, Y.} \& \bibinfo{author}{Wu, J.}
\newblock \bibinfo{title}{{Interpretation of association rules in multi-tier
  structures}}.
\newblock \emph{\bibinfo{journal}{International Journal of Approximate
  Reasoning}} \textbf{\bibinfo{volume}{55}}~(6), \bibinfo{pages}{1439--1457}
  (\bibinfo{year}{2014}).
\newblock \doi{10.1016/j.ijar.2014.04.015} .

\bibitem{Agrawal}
\bibinfo{author}{Agrawal, R.}, \bibinfo{author}{Imieli{\'{n}}ski, T.} \&
  \bibinfo{author}{Swami, A.}
\newblock \bibinfo{title}{{Mining association rules between sets of items in
  large databases}}.
\newblock \emph{\bibinfo{journal}{ACM SIGMOD Record}}
  \textbf{\bibinfo{volume}{22}}~(2), \bibinfo{pages}{207--216}
  (\bibinfo{year}{1993}).
\newblock \doi{10.1145/170036.170072} .

\bibitem{Yazgana2016}
\bibinfo{author}{Yazgana, P.} \& \bibinfo{author}{Kusakci, A.~O.}
\newblock \bibinfo{title}{{A Literature Survey on Association Rule Mining
  Algorithms}}.
\newblock \emph{\bibinfo{journal}{Southeast Europe Journal of Soft Computing}}
  \textbf{\bibinfo{volume}{5}}~(1), \bibinfo{pages}{5--14}
  (\bibinfo{year}{2016}).
\newblock \doi{10.21533/scjournal.v5i1.102} .

\bibitem{Shaukat2017}
\bibinfo{author}{Shaukat~Dar, K.} \& \bibinfo{author}{Zaheer, S.}
\newblock \bibinfo{title}{Association rule mining: An application perspective}.
\newblock \emph{\bibinfo{journal}{International Journal of Computer Science and
  Innovation}} \textbf{\bibinfo{volume}{1}}, \bibinfo{pages}{29--38}
  (\bibinfo{year}{2015}) .

\bibitem{Ghafari2019}
\bibinfo{author}{Ghafari, S.~M.} \& \bibinfo{author}{Tjortjis, C.}
\newblock \bibinfo{title}{A survey on association rules mining using
  heuristics}.
\newblock \emph{\bibinfo{journal}{WIREs Data Mining and Knowledge Discovery}}
  \textbf{\bibinfo{volume}{9}}~(4), \bibinfo{pages}{e1307}
  (\bibinfo{year}{2019}).
\newblock \doi{10.1002/widm.1307} .

\bibitem{7192697}
\bibinfo{author}{Liu, X.} \& \bibinfo{author}{Shen, H.-W.}
\newblock \bibinfo{title}{Association analysis for visual exploration of
  multivariate scientific data sets}.
\newblock \emph{\bibinfo{journal}{IEEE Transactions on Visualization and
  Computer Graphics}} \textbf{\bibinfo{volume}{22}}~(1),
  \bibinfo{pages}{955--964} (\bibinfo{year}{2016}).
\newblock \doi{10.1109/TVCG.2015.2467431} .

\bibitem{Bayardo1999}
\bibinfo{author}{Bayardo, R.~J.} \& \bibinfo{author}{Agrawal, R.}
\newblock \bibinfo{title}{{Mining the most interesting rules}}
  \bibinfo{pages}{145--154} (\bibinfo{year}{1999}).
\newblock \doi{10.1145/312129.312219} .

\bibitem{Brin1997}
\bibinfo{author}{Brin, S.}, \bibinfo{author}{Motwani, R.},
  \bibinfo{author}{Ullman, J.~D.} \& \bibinfo{author}{Tsur, S.}
\newblock \bibinfo{title}{{Dynamic Itemset Counting and Implication Rules for
  Market Basket Data}}.
\newblock \emph{\bibinfo{journal}{SIGMOD Record (ACM Special Interest Group on
  Management of Data)}} \textbf{\bibinfo{volume}{26}}~(2),
  \bibinfo{pages}{255--264} (\bibinfo{year}{1997}).
\newblock \doi{10.1145/253262.253325} .

\bibitem{Bodon2003}
\bibinfo{author}{Bodon, F.} \& \bibinfo{author}{R{\'{o}}nyai, L.}
\newblock \bibinfo{title}{{Trie: An Alternative Data Structure for Data Mining
  Algorithms}} \textbf{\bibinfo{volume}{38}}~(7-9), \bibinfo{pages}{739--751}
  (\bibinfo{year}{2003}).
\newblock \doi{10.1016/0895-7177(03)90058-6} .

\bibitem{Grahne2003a}
\bibinfo{author}{Grahne, G.} \& \bibinfo{author}{Zhu, J.}
\newblock \bibinfo{title}{{Efficiently Using Prefix-trees in Mining Frequent
  Itemsets.}}
\newblock \emph{\bibinfo{journal}{Proc. of the 1st IEEE ICDM Workshop on
  Frequent Itemset Mining Implementations}} \bibinfo{pages}{236--245}
  (\bibinfo{year}{2003}) .

\bibitem{Coenen2004}
\bibinfo{author}{Coenen, F.}, \bibinfo{author}{Leng, P.} \&
  \bibinfo{author}{Ahmed, S.}
\newblock \bibinfo{title}{{Data structure for association rule mining: T-trees
  and P-trees}}.
\newblock \emph{\bibinfo{journal}{IEEE Transactions on Knowledge and Data
  Engineering}} \textbf{\bibinfo{volume}{16}}~(6), \bibinfo{pages}{774--778}
  (\bibinfo{year}{2004}).
\newblock \doi{10.1109/TKDE.2004.8} .

\bibitem{Bui-Thi2021}
\bibinfo{author}{Bui-Thi, D.}, \bibinfo{author}{Meysman, P.} \&
  \bibinfo{author}{Laukens, K.}
\newblock \bibinfo{title}{{MoMAC: Multi-objective optimization to combine
  multiple association rules into an interpretable classification}}.
\newblock \emph{\bibinfo{journal}{Applied Intelligence}}
  (\bibinfo{year}{2021}).
\newblock \doi{10.1007/s10489-021-02595-w} .

\bibitem{Agrawal_apriori}
\bibinfo{author}{Agrawal, R.} \& \bibinfo{author}{S{\&}ant, R.}
\newblock \bibinfo{title}{{Fast Algorithms For Mining Association Rules In
  Datamining}}.
\newblock \emph{\bibinfo{journal}{International Journal of Scientific {\&}
  Technology Research}} \textbf{\bibinfo{volume}{2}}~(12),
  \bibinfo{pages}{13--24} (\bibinfo{year}{2013}) .

\bibitem{Vu2011}
\bibinfo{author}{Vu, L.} \& \bibinfo{author}{Alaghband, G.}
\newblock \bibinfo{title}{{A fast algorithm combining FP-tree and TID-list for
  frequent pattern mining}}.
\newblock \emph{\bibinfo{journal}{Proceedings of Information and Knowledge
  Engineering}} ~(July 2011), \bibinfo{pages}{472--477} (\bibinfo{year}{2011})
  .

\bibitem{Zaki1997}
\bibinfo{author}{Zaki, M.~J.}, \bibinfo{author}{Parthasarathy, S.},
  \bibinfo{author}{Ogihara, M.} \& \bibinfo{author}{Li, W.}
\newblock \bibinfo{title}{{Parallel algorithms for discovery of association
  rules}}.
\newblock \emph{\bibinfo{journal}{Data Mining and Knowledge Discovery}}
  \textbf{\bibinfo{volume}{1}}~(4), \bibinfo{pages}{343--373}
  (\bibinfo{year}{1997}).
\newblock \urlprefix\url{www.aaai.org}.
\newblock \doi{10.1023/A:1009773317876} .

\bibitem{Stancin2019}
\bibinfo{author}{Stancin, I.} \& \bibinfo{author}{Jovic, A.}
\newblock \bibinfo{title}{{An overview and comparison of free Python libraries
  for data mining and big data analysis}}.
\newblock \emph{\bibinfo{journal}{2019 42nd International Convention on
  Information and Communication Technology, Electronics and Microelectronics,
  MIPRO 2019 - Proceedings}} \bibinfo{pages}{977--982} (\bibinfo{year}{2019}).
\newblock \doi{10.23919/MIPRO.2019.8757088} .

\bibitem{Hahsler2023}
\bibinfo{author}{Hahsler, M.}
\newblock \bibinfo{title}{{ARULESPY: Exploring Association Rules and Frequent
  Itemsets in Python}} ~(Raschka 2018) (\bibinfo{year}{2023}).
\newblock \urlprefix\url{http://arxiv.org/abs/2305.15263} .

\bibitem{Yen2019}
\bibinfo{author}{Yen, S.-j.} \& \bibinfo{author}{Chen, A. L.~P.}
\newblock \bibinfo{title}{{A Graph-Based Approach for Discovering Various Types
  of Association Rules}}.
\newblock \emph{\bibinfo{journal}{British Journal of Surgery}}
  \textbf{\bibinfo{volume}{106}}~(Supplement{\_}5), \bibinfo{pages}{11--47}
  (\bibinfo{year}{2019}).
\newblock \doi{10.1002/bjs.11340} .

\bibitem{Koh2010}
\bibinfo{author}{Koh, Y.~S.}, \bibinfo{author}{Pears, R.} \&
  \bibinfo{author}{Yeap, W.}
\newblock \bibinfo{title}{{Valency based weighted association rule mining}}.
\newblock \emph{\bibinfo{journal}{Lecture Notes in Computer Science (including
  subseries Lecture Notes in Artificial Intelligence and Lecture Notes in
  Bioinformatics)}} \textbf{\bibinfo{volume}{6118 LNAI}}~(PART 1),
  \bibinfo{pages}{274--285} (\bibinfo{year}{2010}).
\newblock \doi{10.1007/978-3-642-13657-3{\_}31} .

\bibitem{Hahsler2016}
\bibinfo{author}{Hahsler, M.}
\newblock \bibinfo{title}{{Grouping Association Rules Using Lift}}.
\newblock \emph{\bibinfo{journal}{Proceedings of the 11th INFORMS Workshop on
  Data Mining and Decision Analytics}}  (\bibinfo{year}{2016}).
\newblock \urlprefix\url{http://cran.r-project.org/} .

\bibitem{Jentner2019a}
\bibinfo{author}{Jentner, W.} \& \bibinfo{author}{Keim, D.~A.}
\newblock \bibinfo{title}{{Visualization and Visual Analytic Techniques for
  Patterns}}.
\newblock \emph{\bibinfo{journal}{Studies in Big Data}}
  \textbf{\bibinfo{volume}{51}}, \bibinfo{pages}{303--337}
  (\bibinfo{year}{2019}).
\newblock \urlprefix\url{https://www.eurovis2018.org/}.
\newblock \doi{10.1007/978-3-030-04921-8{\_}12} .

\bibitem{DePadua2018}
\bibinfo{author}{De~Padua, R.}, \bibinfo{author}{Carmo, L. P.~D.},
  \bibinfo{author}{Rezende, S.~O.} \& \bibinfo{author}{De~Carvalho, V.~O.}
\newblock \bibinfo{title}{{An Analysis on Community Detection and Clustering
  Algorithms on the Post-Processing of Association Rules}}.
\newblock \emph{\bibinfo{journal}{Proceedings of the International Joint
  Conference on Neural Networks}} \textbf{\bibinfo{volume}{2018-July}}
  (\bibinfo{year}{2018}).
\newblock \doi{10.1109/IJCNN.2018.8489603} .

\bibitem{Berrado2007}
\bibinfo{author}{Berrado, A.} \& \bibinfo{author}{Runger, G.~C.}
\newblock \bibinfo{title}{{Using metarules to organize and group discovered
  association rules}}.
\newblock \emph{\bibinfo{journal}{Data Mining and Knowledge Discovery}}
  \textbf{\bibinfo{volume}{14}}~(3), \bibinfo{pages}{409--431}
  (\bibinfo{year}{2007}).
\newblock \doi{10.1007/s10618-006-0062-6} .

\bibitem{Buchta2022}
\bibinfo{author}{Hahsler, M.}, \bibinfo{author}{Gr{\"{u}}n, B.} \&
  \bibinfo{author}{Hornik, K.}
\newblock \bibinfo{title}{{Arules - A computational environment for mining
  association rules and frequent item sets}}.
\newblock \emph{\bibinfo{journal}{Journal of Statistical Software}}
  \textbf{\bibinfo{volume}{14}}~(15) (\bibinfo{year}{2005}).
\newblock \doi{10.18637/jss.v014.i15} .

\bibitem{Moahmmed2021}
\bibinfo{author}{Moahmmed, S.~A.}, \bibinfo{author}{Alasow, M.~A.} \&
  \bibinfo{author}{El-Alfy, E. S.~M.}
\newblock \bibinfo{title}{{Clustering of Association Rules for Big Datasets
  using Hadoop MapReduce}}.
\newblock \emph{\bibinfo{journal}{International Journal of Advanced Computer
  Science and Applications}} \textbf{\bibinfo{volume}{12}}~(3),
  \bibinfo{pages}{536--545} (\bibinfo{year}{2021}).
\newblock \doi{10.14569/IJACSA.2021.0120364} .

\bibitem{Li2001}
\bibinfo{author}{Li, W.}, \bibinfo{author}{Han, J.} \& \bibinfo{author}{Pei,
  J.}
\newblock \bibinfo{title}{{CMAR: Accurate and efficient classification based on
  multiple class-association rules}}.
\newblock \emph{\bibinfo{journal}{Proceedings - IEEE International Conference
  on Data Mining, ICDM}} \bibinfo{pages}{369--376} (\bibinfo{year}{2001}).
\newblock \doi{10.1109/icdm.2001.989541} .

\bibitem{AzevedoAClassifiers}
\bibinfo{author}{Azevedo, P.~J.} \& \bibinfo{author}{Rules, C. A.~R.}
\newblock \bibinfo{title}{{A Data Structure to Represent Association Rules
  based Classifiers}}  (\bibinfo{year}{2007}) .

\bibitem{Keim2005}
\bibinfo{author}{Keim, D.~A.}, \bibinfo{author}{Schneidewind, J.} \&
  \bibinfo{author}{Sips, M.}
\newblock \bibinfo{title}{{FP-Viz : Visual Frequent Pattern Mining}}
  (\bibinfo{year}{2005}).
\newblock
  \urlprefix\url{https://mafiadoc.com/fp-viz-visual-frequent-pattern-mining_59e6f7d21723ddf12b1ccc36.html}.

\bibitem{arules:Geng:2006}
\bibinfo{author}{Geng, L.} \& \bibinfo{author}{Hamilton, H.~J.}
\newblock \bibinfo{title}{Interestingness measures for data mining: A survey}.
\newblock \emph{\bibinfo{journal}{ACM Comput. Surv.}}
  \textbf{\bibinfo{volume}{38}}~(3), \bibinfo{pages}{9–es}
  (\bibinfo{year}{2006}).
\newblock \doi{10.1145/1132960.1132963} .

\bibitem{Wu2010}
\bibinfo{author}{Wu, T.}, \bibinfo{author}{Chen, Y.} \& \bibinfo{author}{Han,
  J.}
\newblock \bibinfo{title}{{Re-examination of interestingness measures in
  pattern mining: A unified framework}}.
\newblock \emph{\bibinfo{journal}{Data Mining and Knowledge Discovery}}
  \textbf{\bibinfo{volume}{21}}~(3), \bibinfo{pages}{371--397}
  (\bibinfo{year}{2010}).
\newblock \doi{10.1007/s10618-009-0161-2} .

\bibitem{Luna2018}
\bibinfo{author}{Luna, J.~M.}, \bibinfo{author}{Ondra, M.},
  \bibinfo{author}{Fardoun, H.~M.} \& \bibinfo{author}{Ventura, S.}
\newblock \bibinfo{title}{Optimization of quality measures in association rule
  mining: an empirical study}.
\newblock \emph{\bibinfo{journal}{International Journal of Computational
  Intelligence Systems}} \textbf{\bibinfo{volume}{12}}, \bibinfo{pages}{59--78}
  (\bibinfo{year}{2018}).
\newblock \doi{10.2991/ijcis.2018.25905182} .

\bibitem{Han2004}
\bibinfo{author}{Han, J.}, \bibinfo{author}{Pei, J.}, \bibinfo{author}{Yin, Y.}
  \& \bibinfo{author}{Mao, R.}
\newblock \bibinfo{title}{{Mining frequent patterns without candidate
  generation: A frequent-pattern tree approach}}.
\newblock \emph{\bibinfo{journal}{Data Mining and Knowledge Discovery}}
  \textbf{\bibinfo{volume}{8}}~(1), \bibinfo{pages}{53--87}
  (\bibinfo{year}{2004}).
\newblock \doi{10.1023/B:DAMI.0000005258.31418.83} .

\bibitem{retail}
\bibinfo{title}{{Online Retail}}.
\newblock \bibinfo{howpublished}{UCI Machine Learning Repository}
  (\bibinfo{year}{2015}).
\newblock \bibinfo{note}{Https://doi.org/10.24432/C5BW33}.

\end{thebibliography}
 
\end{document}